\documentclass{article}

\usepackage{arxiv}

\usepackage[utf8]{inputenc} 
\usepackage[T1]{fontenc}    
\usepackage{hyperref}       
\usepackage{url}            
\usepackage{booktabs}       
\usepackage{amsfonts}       
\usepackage{nicefrac}       
\usepackage{microtype}      
\usepackage{lipsum}		
\usepackage{graphicx}
\usepackage{natbib}
\usepackage{doi}
\usepackage{amsmath}

\title{Early Churn Prediction from Large Scale User-Product Interaction Time Series}

\date{July 1, 2022}	

\author{
{Shamik~Bhattacharjee}\\
	Dream11\\
	Mumbai, India\\
	\texttt{shamik.bhattacharjee@dream11.com} \\
	\And
{Utkarsh~Thukral}\\
	Dream11\\
	Mumbai, India\\
	\texttt{utkarsh.thukral@dream11.com} \\
	\And
 {Nilesh~Patil}\\
	Dream11\\
	Mumbai, India\\
	\texttt{nilesh.patil@dream11.com} \\
}

\hypersetup{
pdftitle={A template for the arxiv style},
pdfsubject={q-bio.NC, q-bio.QM},
pdfauthor={David S.~Hippocampus, Elias D.~Striatum},
pdfkeywords={First keyword, Second keyword, More},
}

\begin{document}
\maketitle

\begin{abstract}
Churn, defined as the phenomenon of customers discontinuing their relationship with a business, poses significant economic implications for a variety of Business-to-Customer settings. Success of most system to user actions like promotional discounts, retention campaigns, communications \& recommendations often depends on accurately predicting potential churners as the first goal. This is even more important for fast-moving and transaction heavy domains like fantasy sports, where even regular spending patterns are driven by factors completely out of control ( e.g. high value international sports events affecting smaller parallel events via unexpected secondary and tertiary interactions, user's win/loss in a given contest etc ). Therefore, features that represent transaction history and interaction of a given user with product are good indicators to predict churn, but require deep domain understanding and extensive feature engineering. Feature creation for churn prediction systems also cause large time and resource constraints in production environment where inference pipelines take  $>$ 70\% of total time for feature engineering at the scale of $\sim10^8$ users. This paper presents a comprehensive study focused on predicting customer churn using historical data. Through these analyses, our objective is to develop a model that predicts the likelihood of customer churn, thus enabling businesses to better understand attrition patterns and subsequently devise effective customer retention strategies. We formulate churn prediction as multivariate time series classification and show that predicting churn with user behavior is a powerful method for noisy Business-to-Customer settings. We show that the approach reduces need for extensive feature engineering and outperforms other classical methods prevalent for similar tasks. We present results from our experiments and show that our Transformer based models using limited features, improve upon traditional churn prediction methods.
\end{abstract}


\section{Introduction}
\label{sec:introduction}
The digital revolution has dramatically changed the landscape of sports, and in particular, the growth of fantasy sports has been significant. As of 2023, millions of people participate in online fantasy sports games worldwide, highlighting its transformation from a niche hobby into a mainstream form of entertainment. However, amidst this growing market, retaining customers remains a challenging task, rendering the need to understand and predict customer churn paramount.

Customer churn, defined as the scenario when a customer stops doing business or ends the relationship with a company, represents a considerable risk and cost to any business, especially in non-subscription-based industries like fantasy sports. In any direct Business-to-Customer setting, acquiring new users is typically much more expensive than retaining existing users. High churn rates not only result in lost revenue but also increase the acquisition cost for new customers. In the fiercely competitive fantasy sports industry, where customer loyalty is frequently transient, the ability to predict and proactively address churn is crucial. Retaining existing users requires the ability to know in advance which users are likely to stay and which users are likely to churn out of the platform. With increasing data volumes across multiple touch points and progress of machine learning algorithms in recent years to learn from large datasets, user churn prediction is one of the most direct application areas. 

Churn prediction has emerged as a critical branch of data analysis in many industries, employing statistical modeling, machine learning, and AI techniques to identify the likelihood of customer attrition. This study aims to extend the application of these methods to the fantasy sports industry. By leveraging historical customer data, we seek to create a model that can predict customer churn. This predictive model could offer valuable insights to companies seeking to enhance their customer retention strategies and foster long-term loyalty.

At Dream11, classical ML models for churn prediction are already used at the scale of $~10^8$ users for short future horizon scenarios. The focus now is to not only improve their performance further, but also increase prediction time horizon and reduce complexity of computation required for data pipelines. Some of the classical ML techniques lose out on rich information present in the data by ignoring sequential or contextual information. Modern state-of-the-art techniques in the domain of Natural Language Processing or Computer Vision have shown strong performance in their respective domains since they are backed by automated feature extraction stems which learn how to extract complex features from underlying data given a particular target outcome.

This paper will detail the development and testing of a churn prediction model for the fantasy sports industry. Here, we show that the current state of the art deep-learning techniques, with some architectural modifications, can be directly applied for churn prediction. The goal is to provide a valuable tool for industry stakeholders to mitigate churn, optimize their customer retention strategies, and ultimately, sustain the growth and profitability of their businesses in the highly competitive landscape of fantasy sports.

\section{Related Work}
\label{sec:headings}

The problem of predicting customer churn is a well-studied area in data-science with many different approaches proposed in the literature. These approaches span various industries, from telecommunications and financial services to e-commerce and online gaming. 

Traditional (or classical) machine learning techniques like the Support Vector Machines (SVM) \cite{kim2005application,zhao2005customer,vafeiadis2015comparison,shaaban2012proposed}, Logistic Regression ( LR ) \cite{de2018new,jain2020churn}, Decision Trees ( DT ) \cite{de2018new,hur2005customer} or shallow Artificial Neural Networks (ANN)s have been successful in formulating churn prediction as classification \cite{ahn2020survey}. In \cite{kumar2017optimal}, the authors used a  combination of the support vector machine (SVM) and adaptive boosting (AdaBoost) to classify users with high churn probability. Logistic Regression and Decision Trees are two widely used algorithms for churn prediction, due to their highly interpretability and ease of training. In \cite{de2018new}, the authors used Logistic Regression and Decision Trees to create a hybrid stacked model and the stacked  model shows improved performance compared to individual building block models (LR and DT).

Ensemble algorithms improve upon the performance of constituent learning algorithms. Ensemble techniques combine the prediction outputs of various individual algorithms to provide improved and robust predictions \cite{dong2020survey}. Ensemble algorithms like Bagging \cite{hu2005data} and Boosting \cite{jinbo2007application} have widely been used in churn prediction tasks. Bagging techniques, like the Random Forest classifier \cite{kumar2008predicting, pamina2019effective,ullah2019churn} or its improved variants \cite{xie2009customer} have been successful in churn classification problems. Amongst boosting based algorithms, the XGBoost \cite{pamina2019effective,tang2020customer,ccelik2019comparing} is very successful in this domain.

The literature on the application of deep learning methods for customer churn prediction has traditionally been relatively scarce. However, there has been a recent surge in this research area, indicating a growing recognition of the potential of deep learning in understanding churn behavior. Deep learning models work as powerful feature extractors, that automatically extract highly non-linear features from datasets. This minimizes the need for manual feature engineering. Domingos et al. \cite{domingos2021experimental} applied a Multilayer Perceptron (MLP) and a Deep Neural Network (DNN) on public customer churn data set of a fictitious bank. Cenggoro et al. \cite{cenggoro2021deep} used deep-learning methods for creating embedding vectors which were highly discriminative between potential churning and loyal users. For this task the authors used a publicly available telecommunications dataset of 3333 customers. Their best model achieved a F1 score of 81.16\%.  Panjasuchat and Limpiyakorn (2020) \cite{panjasuchat2020applying} for the first time introduced reinforcement learning for predicting churn. The authors used a Deep-Q Network (DQN) to train on a public telecommunication dataset of $100K$ rows and 99 attributes. To train the DQN agent, the authors used the features vectors as the states and the class labels (churn / no-churn) as the actions. The authors compared their model with other classical-ML models (XGBoost, Random Forest, kNN) and showed that the DQN model was more robust compared with other machine learning models wrt data pattern changes. Cao et al. \cite{cao2019deep} used a Stacked Autoencoder Network (SAE) to extract salient churn-related features from user interaction data. Subsequently, they employed a logistic regression-based classification head for churn classification on the extracted features. In their subscription based service, the authors defined churn as the successive three months of non-subscription window.

In \cite{wangperawong2016churn} , authors represented user interaction data as images and applied Convolutional Neural Networks (CNNs) to predict churn labels based on historical user behavior. The authors evaluated two distinct CNN architectures—DL1 and DL2—which achieved holdout AUC scores of 0.706 and 0.743, respectively. This approach shows a fusion of computer-vision techniques with churn analysis. Furthermore, \cite{fujo2022customer} Fujo et al. undertook an extensive examination of churn prediction within the telecommunication sector. Leveraging a Deep Learning (DL) model on the IBM-Watson public datasets (IBM Telco \& Cell2Cell), they conducted a comprehensive comparative analysis against classical Machine Learning (ML) models. For the IMB Telco dataset, their DL model achieved a 10-fold cross-validation AUC of 86.57\% and a holdout set AUC of 88.11\%. Whereas for the Cell2Cell dataset, the numbers were 73.90\% and 79.38\%. These results were a significant improvement over the classical-ML models applied on the same dataset. This underscores the efficacy of Deep-Learning techniques in enhancing churn prediction accuracy, highlighting its potential in churn prediction studies. In the paper by Umayaparvath et al. \cite{umayaparvathi2017automated}, the authors developed three deep neural network architectures to predict churn on two public datasets, Cell2Cell \& CrowdAnalytix. Performance of their deep-learning based models was as good as traditional classification models.

\section{Formal Problem Formulation} \label{Formal Problem Formulation}



In our study, we aim to predict user churn in a non-subscription based fantasy sports platform. Leveraging the past thirty days of user data, we attempt to predict the churn probabilities for four subsequent weeks. Given a sequence of user activities over the past 30 days, represented as \(X_i = \left[\vec{f}_{t-30}, \vec{f}_{t-29}, \ldots, \vec{f}_{t-1}\right]\), our objective is to predict churn probabilities \(y_1, y_2, y_3, y_4\) for the next four weeks, where \(y_i = 1\) represents churn and \(y_i = 0\) represents no churn in the \(i^{th}\) week.

The aim is to train a model \(y = g(\theta ; x)\) on a training set of user activity sequences and their corresponding churn labels so that the model minimizes the difference between the predicted churn probabilities and the actual churn labels. This can be represented as:

\[ \min_\theta \sum_{i=1}^N \sum_{j=1}^{4} (y_{ij} - g_j(\theta ; X_i))^2 \]

Where, \(N\) is the number of users in the training set, \(y_{ij}\) is the actual churn label for the \(i^{th}\) user in the \(j^{th}\) week and \(g_j(\theta ; X_i)\) is the predicted churn probability for the \(i^{th}\) user in the \(j^{th}\) week.

\subsection{Defining Churn} \label{Defining Churn}


Defining the target variable — or churn — is a critical step, and the definition can vary depending on the user's category or behavior pattern. Here, we are dealing with a situation where churn does not have a uniform definition across all users. Churn in non-contractual systems is defined using consecutive periods of user inactivity. Selection of the exact period depends on the business needs and how user transaction data is distributed. In the absence of explicit indications like subscription cancellations, setting clear and measurable criteria for defining churn is crucial in a non-subscription model. In the context of our study, we have specifically defined churn within the non-subscription based fantasy sports industry as a week of inactivity on our platform. This time frame was chosen as it represents a significant period of inactivity that could indicate a user's decreased engagement or intention to discontinue their involvement in the platform.

This definition of churn – a full week of inactivity – serves as the target variable that our predictive model, built on Transformers and Convolutional Neural Networks, will aim to predict. It’s crucial to note that this definition might not capture all forms of churn, such as users who significantly reduce, but do not entirely cease their activity. However, it provides a practical and measurable churn definition that can be universally applied across the user base for consistent model training and evaluation.

\subsection{\textbf{Reformulation as a Classification Problem :}} \label{Reformulation as a Classification Problem}

In churn prediction, the problem can be framed as a classification task. We aim to predict the churn labels for the next four weeks based on the past 30 days of user vector features. Here, each input and output pair $(X, Y)$ is represented by
\[
([\vec{f}_{t-30}, \vec{f}_{t-29}, \ldots, \vec{f}_{t-1}], [y_{\text{week1}}, y_{\text{week2}}, y_{\text{week3}}, y_{\text{week4}}])
\]

Here, the input features consist of the past 30 days' worth of user vector features $\vec{f}$, and the output vectors represents the output churn labels $y$ for the four future weeks.

Traditionally, churn prediction tasks require extensive feature engineering to extract meaningful representations from raw data. However, deep neural networks have a potential advantage of learning relevant features directly from input data representation, alleviating the need for human knowledge driven feature engineering. This is particularly beneficial in scenarios where complex relationships exist within user activity tracking dataset and output labels.

By leveraging deep learning models, such as convolutional neural networks (CNNs) or transformer models, we can capture intricate dependencies and temporal dynamics within the input sequences. These models can effectively learn and extract high-level representations from the raw data, leading to competitive precision and recall without the need for explicit feature engineering.

Overall, by utilizing deep learning models for churn prediction, we can overcome the limitations of manual feature engineering and achieve competitive accuracy, precision, and recall in the classification task, as the models are adept at learning and extracting meaningful representations from the input data.

\subsection{Modeling Techniques}\label{Modeling Techniques}

Our methodology comprised two distinct approaches, each of which leveraged a different form of user activity data to predict user churn.

\textbf{Approach 1 - Classical Machine Learning}:

In our first approach, we employed Classical Machine Learning models, utilizing aggregated feature vectors to predict the probability of user churn. This approach requires extensive feature engineering since all input features are domain expertise driven and can not be learnt by the learning algorithm.

Let $G_i$ represent the feature vector for user $i$ that we've constructed from aggregated data. Our Classical Machine Learning models then predict future churn probabilities $P(C_i | G_i)$ for user $i$ over the next four weeks. This can be written as follows:
\begin{equation}
P(C_i | G_i) = Model(G_i)    
\end{equation}
where $Model$ refers to our Classical Machine Learning models (Logistic Regression, Random Forest, or Gradient Boosting Trees), $G_i$ is the feature vector for user $i$, and $C_i$ is the churn label vector for user $i$ for the next four weeks.

\textbf{Approach 2 - Deep Learning Models}: 

The second approach involved Deep Learning models, specifically Convolutional Neural Networks (CNNs) and attention based models which were trained on the users' past 30 days of time-series feature data to predict churn. This can be mathematically represented as:

Let $F_i = = \left[\vec{f}_{t-30}, \vec{f}_{t-29}, \ldots, \vec{f}_{t-1}\right]$ represent the sequence of 30-day feature vectors for user $i$ that we've constructed from daily transaction data. Our Deep Learning models then predict the churn probabilities $P(C_i | F_i)$ for user $i$ over the next four weeks. This can be written as follows:
\begin{equation}
P(C_i | F_i) = Model(F_i)    
\end{equation}

Here, Model refers to our Deep Learning models: Vanilla CNN, VGG-like \cite{krizhevsky2012imagenet}, CNN(filter-width=n$_{features}$), CNN(filter-height=$\tau$), Inception-Resnet \cite{szegedy2015going, szegedy2016rethinking, szegedy2017inception}, ConvNeXt \cite{li2022convnext}, Transformer \cite{vaswani2017attention, liu2021swin}

The comparative analysis of these two approaches facilitates the evaluation of the impact of feature representation and model complexity on the prediction of user churn.

Our quest to improve predictive performance in the task of customer churn prediction led us to explore a wide array of deep learning models. These models, each with their own unique capabilities for processing and learning from time-series data, ranged from simpler architectures like Convolutional Neural Networks (CNNs) to more intricate designs such as custom Inception-ResNet hybrids and Transformer Encoders.

We commenced our journey with the traditional CNN \cite{krizhevsky2012imagenet, simonyan2014very, szegedy2015going, he2016deep}, a popular choice for time-series analysis due to its capacity to discern patterns over fixed-sized windows. Convolution neural networks have been extensively used in problems involving time-series forecasting or classification tasks \cite{cui2016multi, chen2015convolutional, ismail2019deep}. Time-series classification or forecasting has wide range of applications ranging from stock market prediction \cite{hoseinzade2019cnnpred}, weather forecasting \cite{weyn2020improving}, disease detection \cite{dobko2020cnn}, to customer churn prediction. We benchmark with vanilla CNN ( \textit{VGG-like} ) for our classification task involving the multivariate user time-series data. As we sought to enhance our model's performance, we extended the vanilla CNN in two distinct ways: First, we modified the kernel width to align with the dimension of the feature vectors, enabling our model to detect interactions among all features within a given day. Second, we adjusted the kernel height to span the entirety of the 30-day time window, thus allowing our model to capture long-term temporal patterns within the user activity data.

To obtain even higher predictive performance, We developed a custom hybrid of Inception \cite{szegedy2017inception} and ResNet \cite{he2016deep} models, designed to extract multi-scale feature patterns while addressing the vanishing gradient problem common in deep networks \cite{szegedy2017inception,he2016deep}. Variations of the Inception-ResNet architecture has also been used for time-series classification tasks. We explore the \textit{InceptionTime} \cite{ismail2020inceptiontime} architecture for our use case. However, we observe a deterioration in performance. 

Following this, we implemented a ConvNeXt architecture \cite{liu2022convnet}. ConvNeXt is renowned for its ability to capture intricate feature interactions and reduce computational overhead through its unique grouped convolution design, making it an attractive choice for our task. In the present work, we used a scaled-down version of the \textit{ConvNeXt} model for the classification task on our reshaped two-dimensional multivariate time-series data.

Finally, we utilized a variant of the Transformer architecture \cite{vaswani2017attention}, specifically a scaled-down version of the Vision Transformer (ViT) model proposed by Dosovitskiy et al \cite{dosovitskiy2020image}. This version retained the self-attention mechanism of the original model, allowing it to focus on different parts of the input sequence and recognize intricate temporal patterns across the 30-day user activity window. By only using the encoder mechanism, we could efficiently handle time-series data, while reducing the complexity introduced by the decoder mechanism in the full transformer model.

This broad spectrum of deep learning models provided us with an array of tools to understand and predict user churn. Each model, with its unique strengths and features, lent different insights into the user activity data. In the following sections, we will delve into a more detailed comparison of these models, and discuss the implications of our findings.

\section{Data Preparation}\label{Data Preparation}

The process of preparing the dataset for our predictive models was a meticulous one, encompassing several crucial stages. Our primary objective was to construct a detailed yet comprehensible representation of each user's interaction history that could be effectively processed by both classical Machine Learning and Deep Learning models. Given the scale required for preparing this dataset, we split data
preparation into 2 stages :
(1) \textbf{Level-01} Features from transaction level data
(2) \textbf{Level-02} Features from \textbf{Level-01} features.
The raw data form is shown in Table \ref{tab:raw_transactional_data} and we use these transactions data to create \textbf{Level-01} features.

Representing user transactional data in a table, we might have a structure similar to the following:

\begin{table}[h]
\centering
\begin{tabular}{c|c|c|c}
\textbf{Transaction ID} & \textbf{User ID} & \textbf{Timestamp} & \textbf{Feature Vector} \\ \hline
$1$ & 1 & $t_{1,1}$ & $\vec{f}_{1,1}$ \\ 
$2$ & 1 & $t_{2,1}$ & $\vec{f}_{2,1}$ \\ 
\vdots & \vdots & \vdots & \vdots \\
$n_1$ & 1 & $t_{n_1,1}$ & $\vec{f}_{n_1,1}$ \\ \hline
\vdots & \vdots & \vdots & \vdots \\ \hline
$N$ & $M$ & $t_{1,M}$ & $\vec{f}_{1,M}$ \\ 
$N+1$ & $M$ & $t_{2,M}$ & $\vec{f}_{2,M}$ \\ 
\vdots & \vdots & \vdots & \vdots \\
$N+n_M$ & $M$ & $t_{n_M,M}$ & $\vec{f}_{n_M,M}$ \\ 
\end{tabular}
\caption{Example of raw transactional data}
\label{tab:raw_transactional_data}
\end{table}

In Table \ref{tab:raw_transactional_data}, each row is representative of a unique transaction, identified distinctly by a \textbf{Transaction ID}. The \textbf{User ID} field pertains to the specific user who was involved in the transaction, while the \textbf{Timestamp} field indicates the precise time at which the transaction occurred. The array of features tied to each transaction is denoted in the \textbf{Feature Vector} column. These transactional data sets form the cornerstone for generating the Level-1 features that will further be used in the modeling process.

This raw transaction data is then aggregated on a day level to form the \textbf{Level-01} features that feed into the model (Table \ref{tab:aggregated_data}). The \textbf{Level-01} features will be vectors of aggregated feature values (such as count, sum, mean, etc.) for each user for each day.

Here's how the \textbf{Level-01} features can be represented for each user:

\begin{table}[h]
\centering
\begin{tabular}{c|c|c|c}
User ID & Date & Feature Vector & No. of Transactions \\
\hline
1 & $d_{1,1}$ & $\vec{f}_{1,1}$ & $n_{1,1}$ \\
\cline{2-4}
& $d_{1,2}$ & $\vec{f}_{1,2}$ & $n_{1,2}$ \\
\cline{2-4}
& $\vdots$ & $\vdots$ & $\vdots$ \\
\cline{2-4}
& $d_{1,D_1}$ & $\vec{f}_{1,D_1}$ & $n_{1,D_1}$ \\
\hline
2 & $d_{2,1}$ & $\vec{f}_{2,1}$ & $n_{2,1}$ \\
\cline{2-4}
& $d_{2,2}$ & $\vec{f}_{2,2}$ & $n_{2,2}$ \\
\cline{2-4}
& $\vdots$ & $\vdots$ & $\vdots$ \\
\cline{2-4}
& $d_{2,D_2}$ & $\vec{f}_{2,D_2}$ & $n_{2,D_2}$ \\
\hline
$\vdots$ & $\vdots$ & $\vdots$ & $\vdots$ \\
\hline
$M$ & $d_{M,1}$ & $\vec{f}_{M,1}$ & $n_{M,1}$ \\
\cline{2-4}
& $d_{M,2}$ & $\vec{f}_{M,2}$ & $n_{M,2}$ \\
\cline{2-4}
& $\vdots$ & $\vdots$ & $\vdots$ \\
\cline{2-4}
& $d_{M,D_M}$ & $\vec{f}_{M,D_M}$ & $n_{M,D_M}$ \\
\end{tabular}
\caption{Level-1 User Feature Table. Each row under a user ID represents a day's aggregated features of that user.}
\label{tab:aggregated_data}
\end{table}

Using the \textbf{Level-01} aggregates, we create the training data. For each user, we consider the past $30$ days of data as the time series input and generate the corresponding churn label for the next four weeks. The churn label for each week indicates whether the user has churned (inactive) or not.

To clarify this in the context of the table, we can define the training data as follows:

- Input (X): For each user, we extract the feature vectors from the past 30 days, denoted as $\vec{f}_{t-30}, \vec{f}_{t-29}, \ldots, \vec{f}_{t-1}$, where $t$ represents the timestamp of the last transaction in the training data.

- Output (Y): The churn label for each week is represented as a binary value, indicating whether the user has churned (1) or not churned (0) in that specific week. Let's denote these labels as $y_{\text{week1}}, y_{\text{week2}}, y_{\text{week3}}, y_{\text{week4}}$.

Hence, the training data can be represented as a pair of input-output sequences:

\[
\left(\left[\vec{f}_{t-30}, \vec{f}_{t-29}, \ldots, \vec{f}_{t-1}\right], \left[y_{\text{week1}}, y_{\text{week2}}, y_{\text{week3}}, y_{\text{week4}}\right]\right)
\]

This training data can then be used to train models for churn prediction using various techniques, such as classical machine learning models or deep learning models.

In the next stage, we leveraged the \textbf{Level 01} features to create \textbf{Level-02} data by conducting various aggregations such as mean ($\mu$), standard deviation ($\sigma$), and other statistical measures over the 30-day window. This higher-level aggregation aimed to capture different aspects of the user's behavior, including their consistency and variability in their activity over time. Formally, for a user $i$ and a feature $j$, the \textbf{Level-02} features were defined as:

\[\mu_{i,j} = \frac{1}{30}\sum_{k=0}^{d} f_{i,k,j}\]
\[\sigma_{i,j} = \sqrt{\frac{1}{30}\sum_{k=0}^{d} (f_{i,k,j} - \mu_{i,j})^2}\]

The resulting \textbf{Level-02} data offered a more condensed and generalized view of the user's historical activity, which was particularly suited for our classical Machine Learning models. The \textbf{Level-02} feature vector $G_i$ for user $i$ was then formed as:

\[G_i = \{\mu_{i,1}, \sigma_{i,1}, ..., \mu_{i,N}, \sigma_{i,N}\}\]



Since our model is designed to predict future user churn, we construct a binary target vector $Y_i$ for each $user_i$, where each element $y_{i,w}$ of the vector corresponds to whether the $i^{th}$ user churns in week $w$ (1 to 4) after the end of the 30-day activity window:

\[Y_i = \{y_{i,1}, y_{i,2}, y_{i,3}, y_{i,4}\}\]

Formally, for each week $w_j$ after the 30-day window, $y_{i,w}$ is defined as:

\[y_{i,w} = 
\begin{cases} 
1 & \text{{if user i has no activity in week } w} \\
0 & \text{otherwise}
\end{cases}\]

It is important to note that by designing the target as a vector, we enable our model to provide a more granular prediction, indicating not just if a user will churn, but also when they are likely to churn within a four-week horizon. This additional information can be of significant value to various business operations, such as customer engagement and retention strategies.

For our current work, we selected eleven features from a pool of a large number of available features. These eleven features were shortlisted through rigorous exploratory data analysis (EDA) and domain knowledge.

\subsection{Data Sampling}\label{Data Sampling}
We run our experiments on sampled data from Dream11's fantasy sports user base. We use user transaction history records between 2018-01-01 and 2020-12-31. The setting here is non-contractual churn. We first create a random sample of Dream11 users and process their transactions for creating modeling dataset for our experiments. This modeling dataset has $\sim 10^6$ distinct users and $\sim 10^8$ raw transactions with various different types of transactions. This large dataset provides a comprehensive historical perspective of user interactions with our fantasy sports platform. Using this dataset, we created the train, validation and test datasets. The \textit{Train}, \textit{Validation} and \textit{Test} ratio was chosen to be $0.75$, $0.05$ and $0.20$ respectively.

Here, We are not mentioning individual sporting events because we are not doing analysis for individual sporting events. We are considering users' past history irrespective of individual sporting events across all models. Any such system-level features can be included in the same formulation by adding additional covariates.

\section{Training Setup}\label{Training Setup}

Leveraging this vast dataset, our deep learning models were trained using the PyTorch framework, a popular open-source machine learning library for Python. We  follow the same training protocols for all models. All models were initialized with random weights and trained for 100 epochs. In brief, we used the Adam optimizer \cite{kingma2014adam} to train our models.

In terms of computational setup, our training process was conducted in a distributed manner across multiple GPUs, ensuring efficient handling of our large-scale data and complex models. We utilized Horovod, a distributed deep learning training framework, to coordinate the training process across these GPUs. Horovod facilitates synchronous distributed training, enabling each participating GPU to independently compute the gradients for the subset of data it was allocated and then collectively averaging these gradients to update the model weights. This method ensures model consistency and accelerates the training process by allowing concurrent computations.

The process of loading data into our training environment was facilitated by Petastorm, an open-source data access library. It enabled efficient and high-speed loading of our large-scale data directly from Amazon S3 into our PyTorch models. Petastorm creates a streamlined interface between the data stored in S3 and the PyTorch DataLoader, simplifying the data ingestion process and enhancing the performance of our training operations.

Given the complex nature of our models and the size of our data, the training process was set to run for 100 epochs. This balance was struck to allow sufficient learning time for our models to capture intricate patterns in the data while avoiding excessive training that could lead to overfitting. Over the course of these epochs, the model performance was continuously monitored, and adjustments were made as necessary to optimize the prediction accuracy.


We conducted grid search over all the tunable hyper-parameters on the held-out validation set for each of the models. The global batch size was chosen $16$K training samples. During training, each GPU was allocated $\sim 4$K training samples. This batch size was carefully chosen to achieve best training performance at the shortest possible training time. The learning-rate parameter ($lr$) for all the models was kept fixed at $10^{-4}$. The number of trainable parameters for different models varied significantly and hence total training time also differed for each. Our custom transformer architecture has approximately $1.8$ million trainable parameters, featuring a composition of eight self-attention blocks. The custom Inception-Resnet comprises nine inception blocks, each encompassing convolution operations with three different kernel sizes: $ 3 \times 3 $, $ 5 \times 5$ and $ 7 \times 7 $. The architecture contained a total of approximately $2.4$ million trainable parameters. Our adapted ConvNeXt architecture featured three convolution layers with approximately $0.8$ million trainable parameters. For our Long Short-Term Memory (LSTM) model, we designed a five-layer architecture, followed by a classification head comprised of two dense layers. The total count of trainable parameters here was approximately one million. Both the Convolutional Neural Network architectures have approximately one million trainable parameters. Here also a two layer dense neural network has been used as the classification head for predicting the churn label.

For all models, we used a Batch-Normalization \cite{ioffe2015batch} layer to improve stability and ensure faster training convergence. To reduce model over-fitting, Dropout layers \cite{srivastava2014dropout} were used. The Dropout probability ($p$) was chosen carefully for the individual models and it varied from $0.1$ to $0.4$. Training time for CNN was least (~6 hours) while training a transformer model took the longest (~13 hours). We experimented with N-epochs = [10, 20, 50, 75, 100] and observed that model metrics improvement slowed down after 20 epochs and did not benefit from any additional epochs after 50.

4 Nvidia-A10 GPUs were used for training Deep learning models while Spark was used for creating features at different aggregation levels over the complete dataset and instrumenting distributed training using Horovod \cite{sergeev2018horovod} and Petastorm .

\section{Results}{\label{Results}}

A receiver operating characteristic curve, or $ROC\ curve$, is used to measure model performance of a binary classifier as the discrimination threshold on top of predicted scores is varied. We calculate True Positive Rate ($TPR$) and False Positive Rate ($FPR$) for each threshold $\in [0, 1]$ as shown below.

\begin{equation}
TPR = \frac{TP}{ TP + FN}
\end{equation}
\begin{equation}
FPR = \frac{FP}{ FP + TN}
\end{equation}

\begin{figure}
\centering
    \begin{tabular}{c c c c}
        \includegraphics[width = 0.25\linewidth, height=0.25\linewidth]{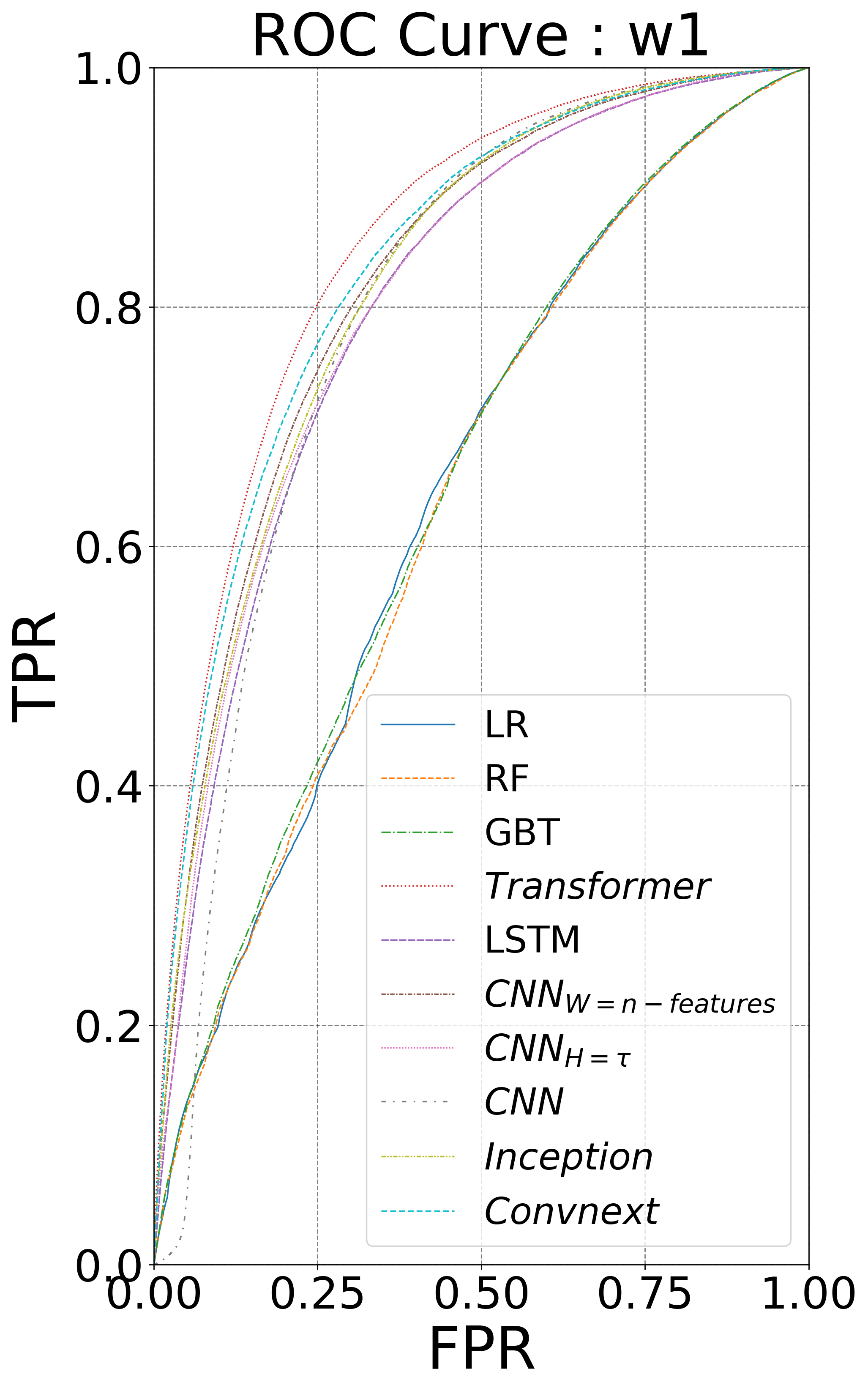} & 
        \includegraphics[width = 0.25\linewidth, height=0.25\linewidth]{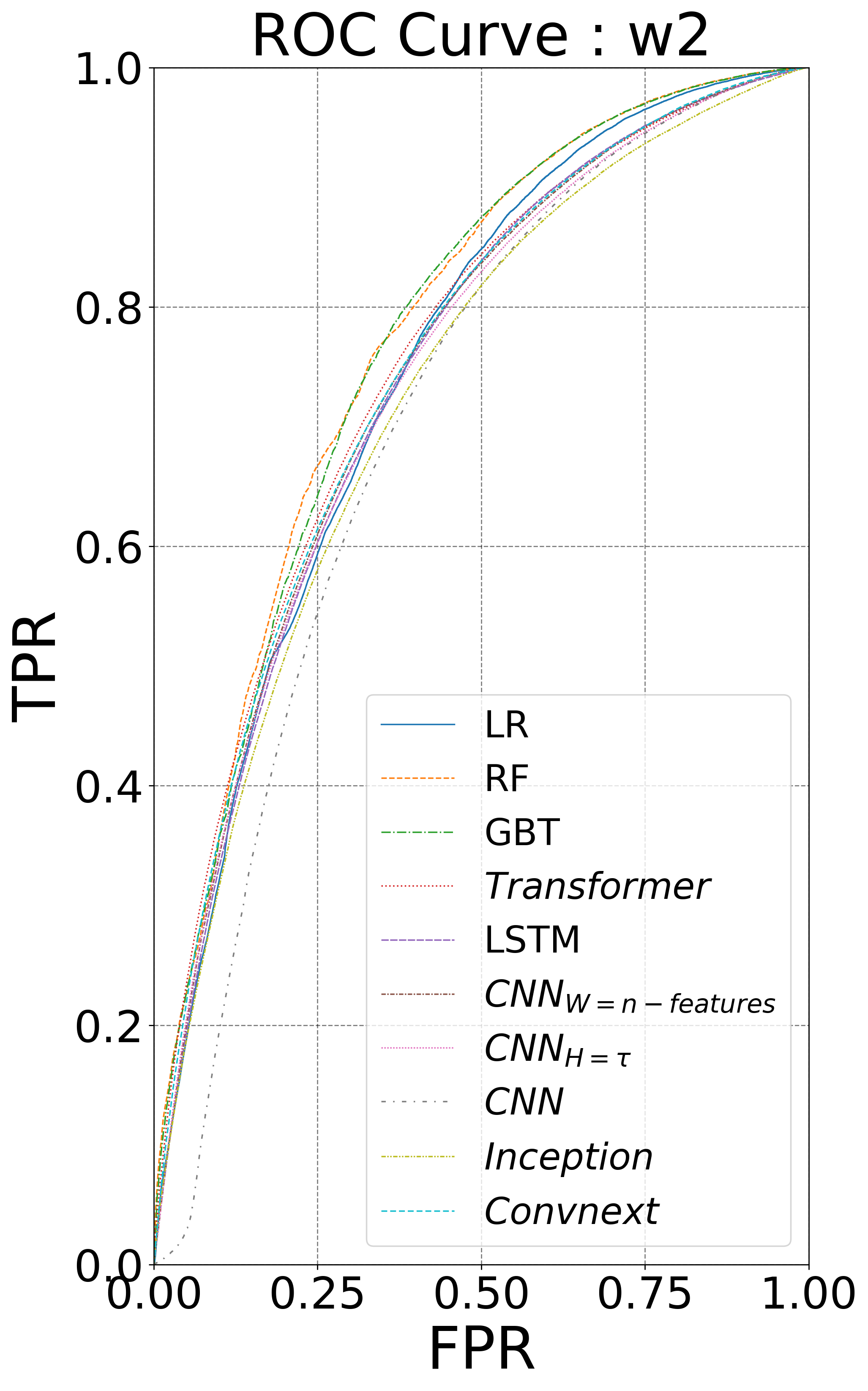} &
        \includegraphics[width = 0.25\linewidth, height=0.25\linewidth]{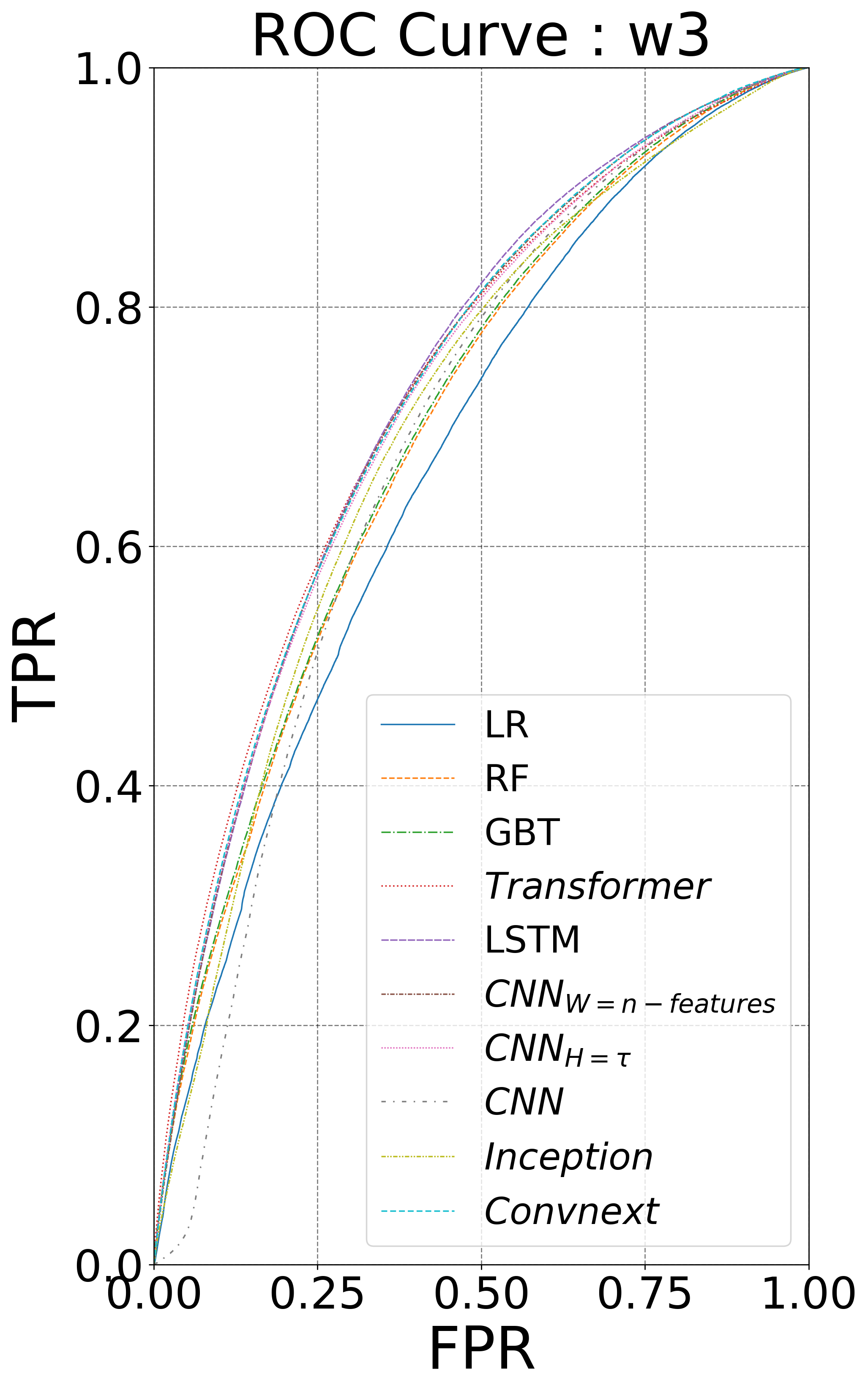} & 
        \includegraphics[width = 0.25\linewidth, height=0.25\linewidth]{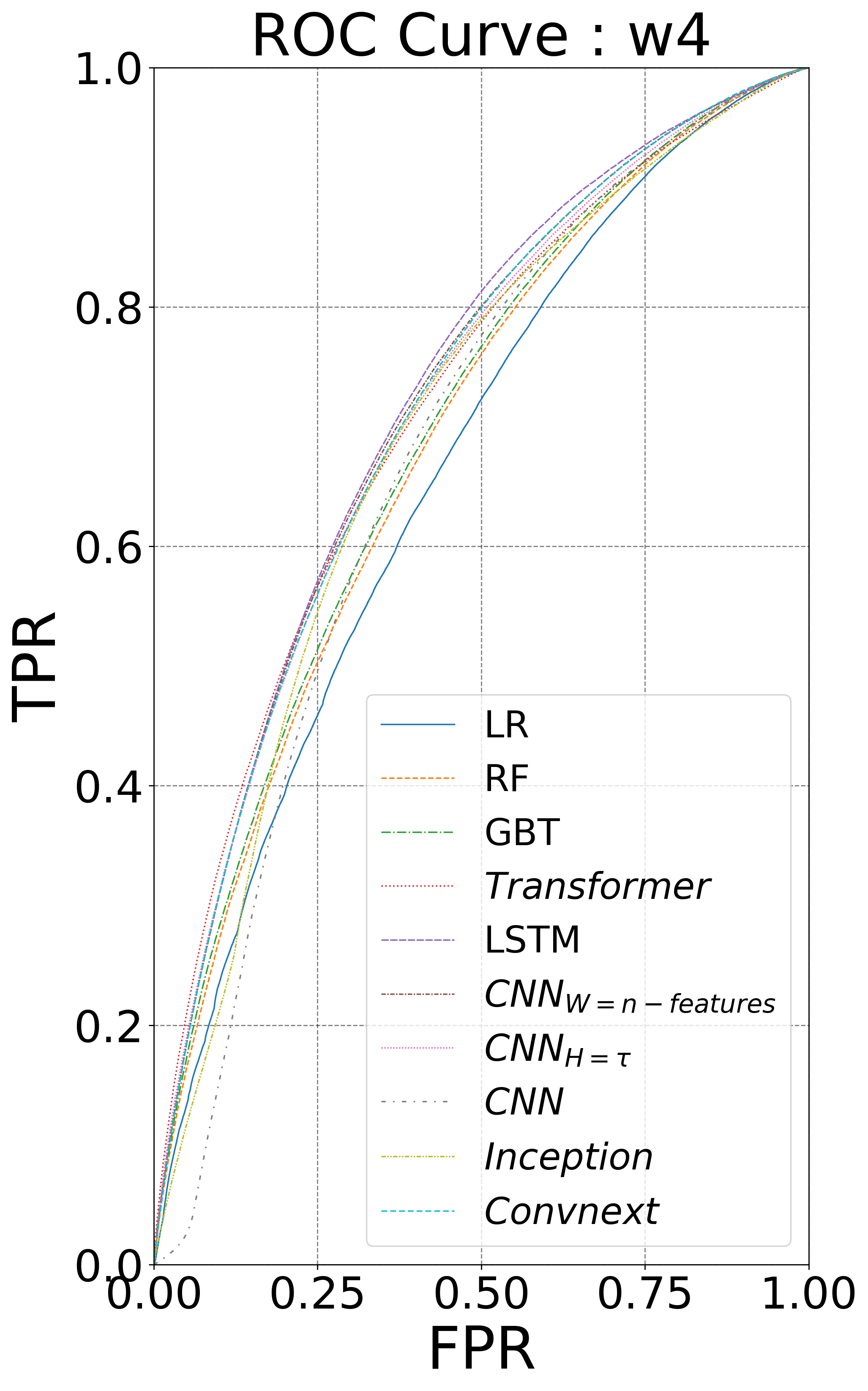} \\
        \includegraphics[width = 0.25\linewidth, height=0.25\linewidth]{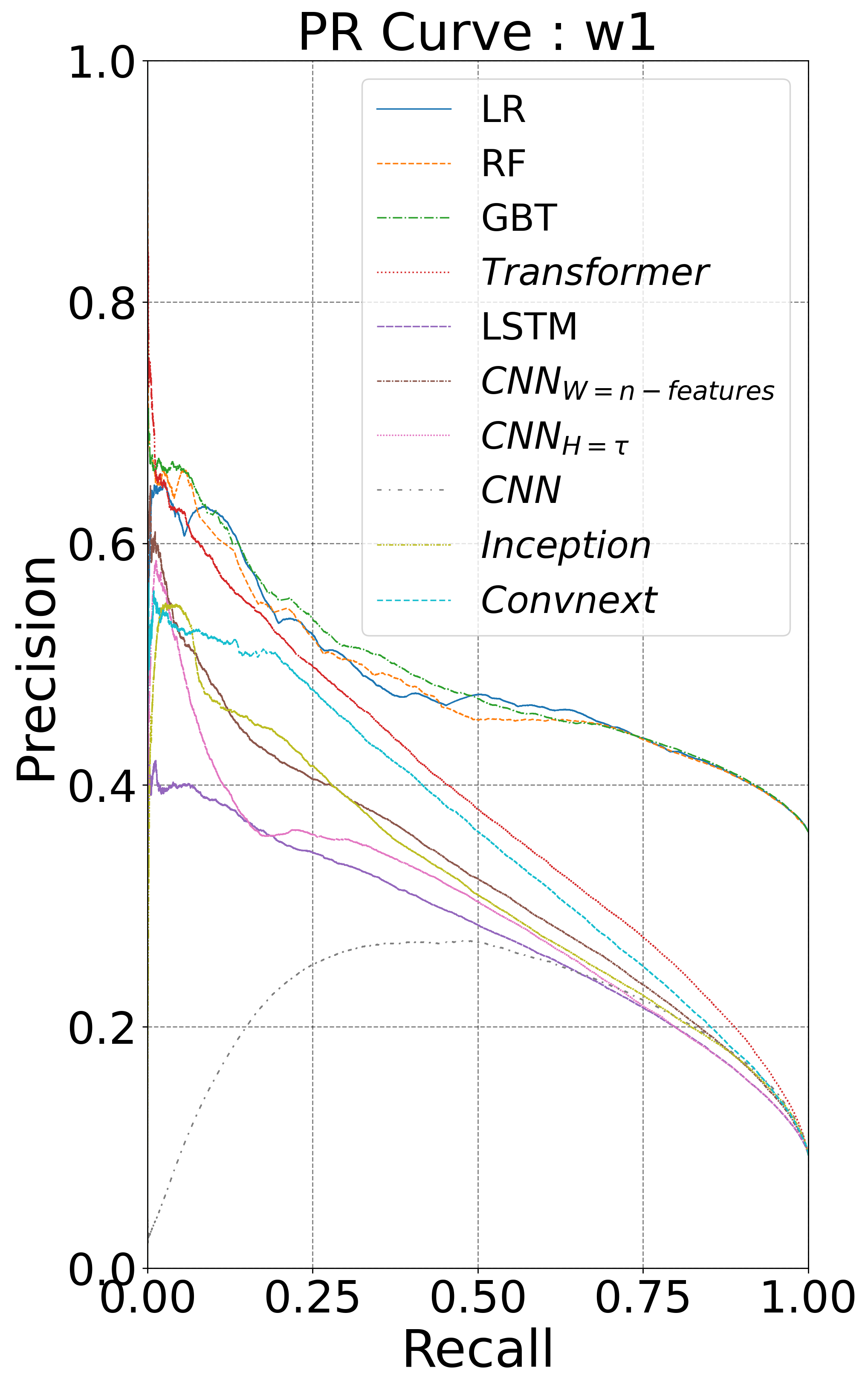} & 
        \includegraphics[width = 0.25\linewidth, height=0.25\linewidth]{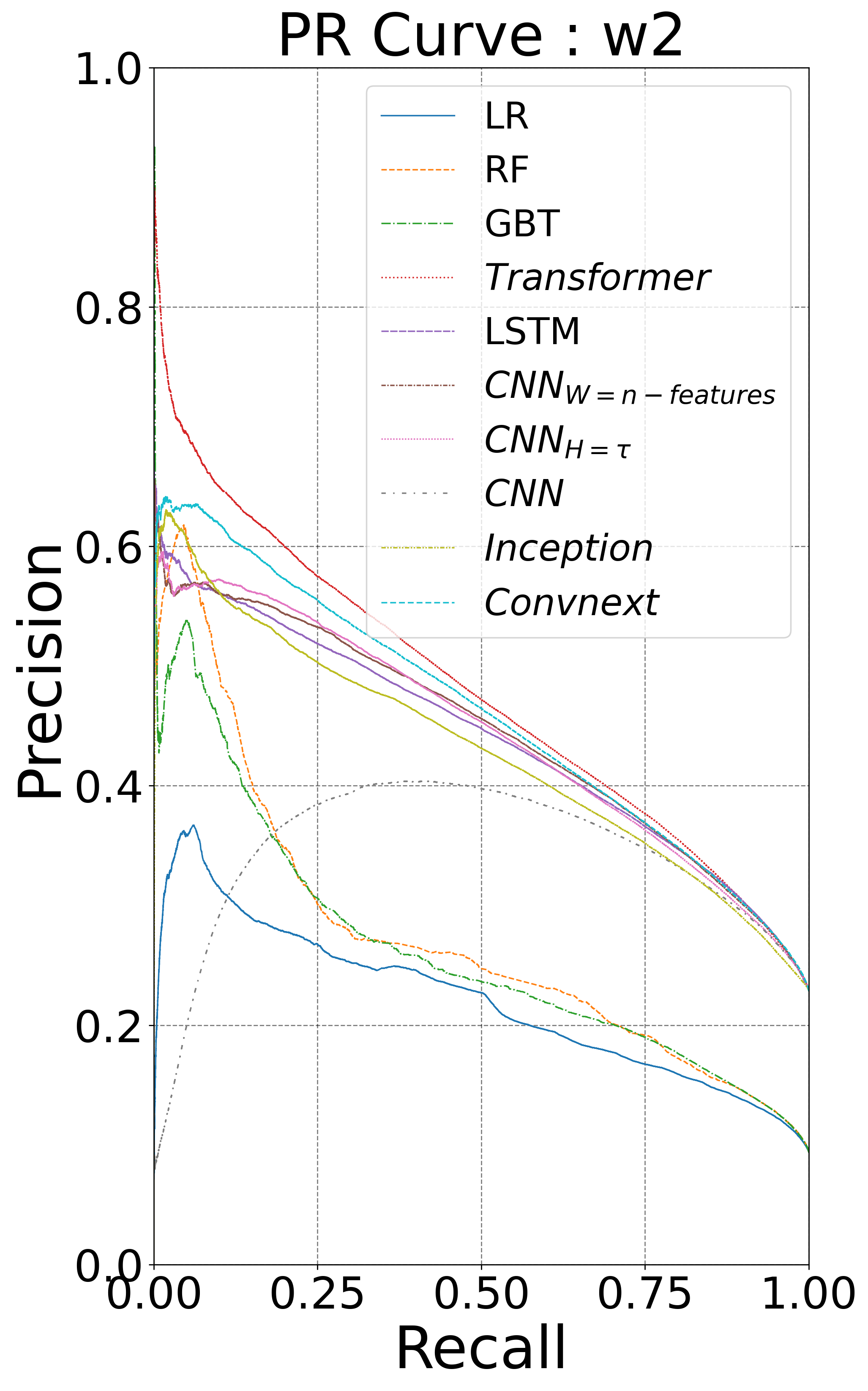} &
        \includegraphics[width = 0.25\linewidth, height=0.25\linewidth]{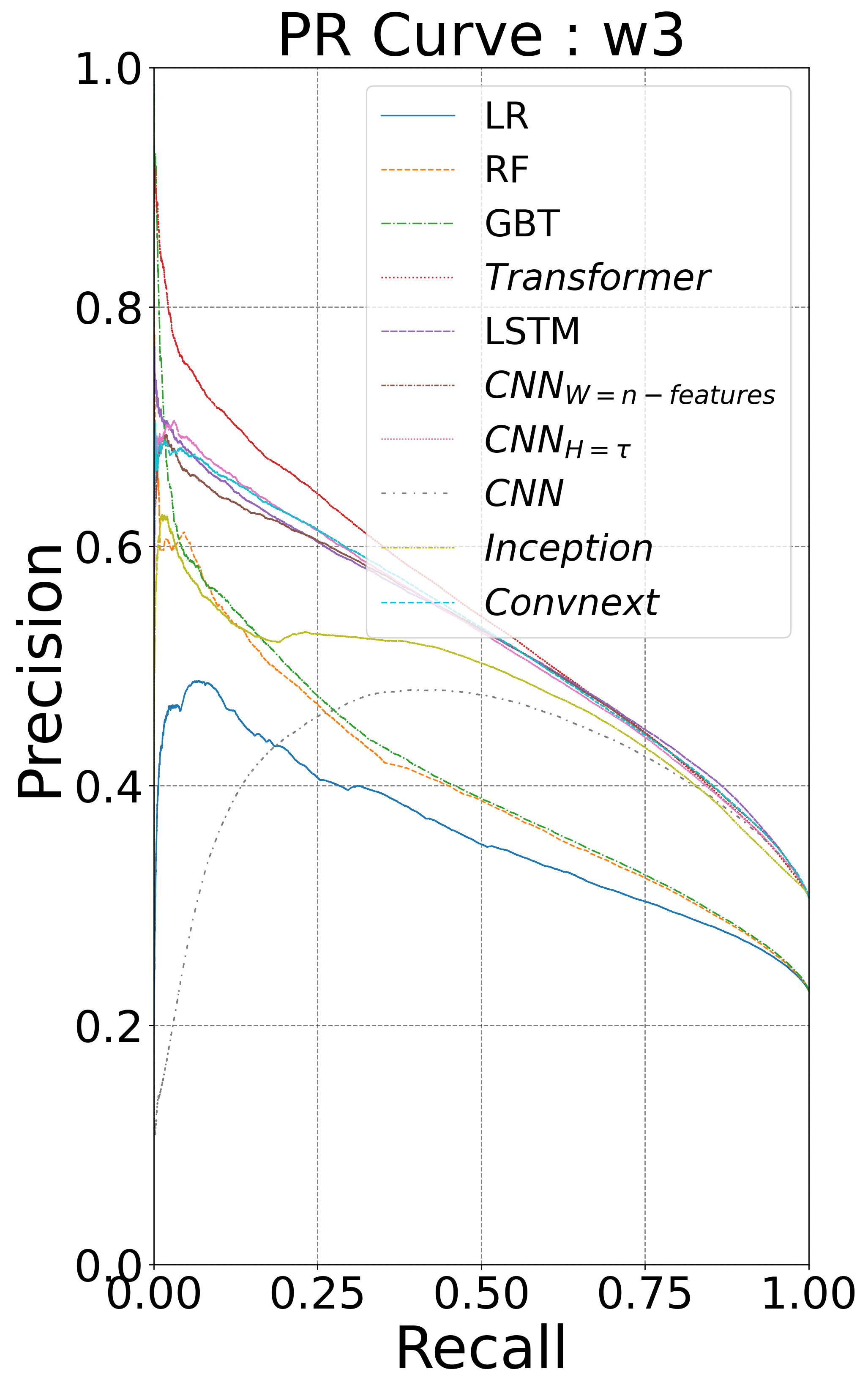} & 
        \includegraphics[width = 0.25\linewidth, height=0.25\linewidth]{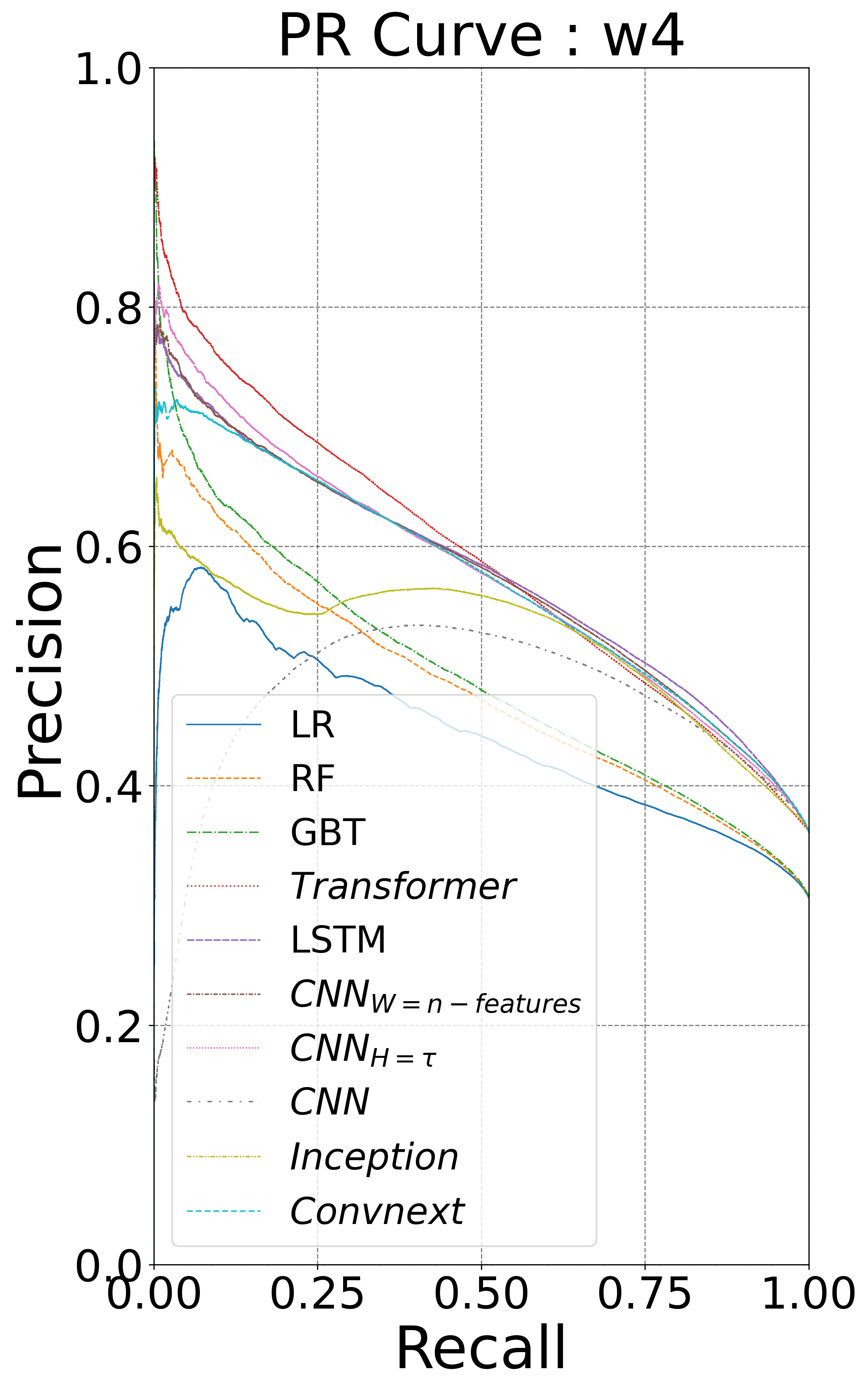}
    \end{tabular}
\caption{Model Performance}
\label{Model Performance}
\end{figure}

A high level summary of results from all of our experiments is presented in Table \ref{tab:ModelPerformance}. Our best performing model was the transformer architecture which shows \textbf{ $\sim \textbf{6\%}$ improvement on an average across all time periods} compared to best performing GBT model. Complete $ROC$ Curves for each model are also presented (Fig. \ref{Model Performance}) which are used to calculate the $AUC$ benchmark.

We also present $Precision-Recall$ Curves for all of our models (\ref{Model Performance}), showing substantial difference between each model variant. In our use case, there is always a trade-off involved between high false $+ve$ rate vs retention budgets. The goal is to get as many as users as we can without wasting treatment budgets. This trade-off has different thresholds for each application, e.g. for giving promotional discounts using predicted scores, we have to be more stringent while for communication related actions, we have substantial leeway.

\begin{table}[h]
\centering
\begin{tabular}{c c c c c}
 \hline
 Classifier & $AUC_{W_{01}}$ & $AUC_{W_{02}}$ & $AUC_{W_{03}}$ & $AUC_{W_{04}}$ \\
 \hline\hline
 LR          & 0.644 & 0.749 & 0.672 & 0.662 \\ 
 RF          & 0.640 & 0.775 & 0.699 & 0.687 \\ 
 GBT         & 0.646 & 0.773 & 0.703 & 0.694 \\ 
 CNN         & 0.797 & 0.705 & 0.686 & 0.675 \\
 ConvNext    & 0.835 & 0.751 & 0.729 & 0.717 \\ 
 Inception-Resnet   & 0.819 & 0.729 & 0.704 & 0.696 \\ 
 $CNN_{H=\tau}$ & 0.807 & 0.742 & 0.725 & 0.716 \\ 
 $CNN_{W=N}$ & 0.822 & 0.747 & 0.728 & 0.729 \\ 
 LSTM    & 0.800 & 0.742 & 0.727 & 0.721 \\ 
 Transformer & \textbf{0.858} & \textbf{0.756} & \textbf{0.732} & \textbf{0.716} \\ 
 \hline
\end{tabular}
\caption{Model Performance}
\label{tab:ModelPerformance}
\end{table}


\subsection{Results from Classical Models : }\label{Results from Classical Models}

As shown  in Table \ref{tab:ModelPerformance}, results from Logistic Regression are worse across all time periods (i.e. Week - 01 / 02 / 03 / 04 churn ) but it is a sensible benchmark against which other modeling approaches can be tested.
Ensemble models perform better compared to the LR model. The PR Curve for GBT vs LR clearly shows performance improvement as we move from LR $\longrightarrow$ GBT. Overall, these methods under-perform substantially compared to Transformer based model as shown in Table \ref{tab:ModelPerformance}.

\subsection{Results from Deep Networks : }\label{Results from Deep Networks}

Our Transformer model has consistently higher performance compared to others with an additional training time of ~$15\%$ for the current version. Based on ROC and PR curves in Model Performance, there are certain instances where ROC curve for classical algorithms is better compared to Deep learning based models. Taking a deeper look through the PR curves though, shows substantially poor performance. This is primarily caused by data imbalance in churn problem statements, where label distribution is heavily skewed towards one label. In our case, different weeks have different skews ranging from $2$x to $10$x which shows up in poor performance of models with lower learning capacity.

\section{Conclusions and Summary}\label{Conclusions and Summary}

In the current paper, we presented our work on User Churn Prediction. We first formulated this problem as a classification problem statement and proposed two distinct approaches to solve it for real world datasets. In the first approach, we aggregated a user's activity time series into a row vector and build a classifier on top of this representation. To improve performance of this approach, extensive domain knowledge is required to create new features from a user's time series. With this approach, results from GBTs perform better compared to other classical ML methods.

Second approach creates a 2D array of chronologically arranged observations ($X_{k,\tau}$) as input for the DL models. Here, we used DL models which take advantage of the sequential nature as well as interaction between the features. The highest performing model was the Transformer architecture which outperforms prediction by the classical models by $\sim 6\%$ on an average. With this improved approach, there's a substantial scope for exploring seq-2-seq applications in other time series applications ( e.g. CLTV prediction, Forecasting, sequential product recommendations being some of the top use-cases ).

A major limitation of this approach is requirement of large training data volume and long training duration. The first limitation is usually less relevant for forecasting user metrics - given high user counts in most Business to Customer settings, but could prove to be a problem for other entities used in forecasting.

\bibliographystyle{unsrtnat}
\bibliography{references}  

\begin{thebibliography}{47}
\providecommand{\natexlab}[1]{#1}
\providecommand{\url}[1]{\texttt{#1}}
\expandafter\ifx\csname urlstyle\endcsname\relax
  \providecommand{\doi}[1]{doi: #1}\else
  \providecommand{\doi}{doi: \begingroup \urlstyle{rm}\Url}\fi

\bibitem[Kim et~al.(2005)Kim, Shin, and Park]{kim2005application}
Sun Kim, Kyung-shik Shin, and Kyungdo Park.
\newblock An application of support vector machines for customer churn analysis: Credit card case.
\newblock In \emph{International Conference on Natural Computation}, pages 636--647. Springer, 2005.

\bibitem[Zhao et~al.(2005)Zhao, Li, Li, Liu, and Ren]{zhao2005customer}
Yu~Zhao, Bing Li, Xiu Li, Wenhuang Liu, and Shouju Ren.
\newblock Customer churn prediction using improved one-class support vector machine.
\newblock In \emph{International conference on advanced data mining and applications}, pages 300--306. Springer, 2005.

\bibitem[Vafeiadis et~al.(2015)Vafeiadis, Diamantaras, Sarigiannidis, and Chatzisavvas]{vafeiadis2015comparison}
Thanasis Vafeiadis, Konstantinos~I Diamantaras, George Sarigiannidis, and K~Ch Chatzisavvas.
\newblock A comparison of machine learning techniques for customer churn prediction.
\newblock \emph{Simulation Modelling Practice and Theory}, 55:\penalty0 1--9, 2015.

\bibitem[Shaaban et~al.(2012)Shaaban, Helmy, Khedr, and Nasr]{shaaban2012proposed}
Essam Shaaban, Yehia Helmy, Ayman Khedr, and Mona Nasr.
\newblock A proposed churn prediction model.
\newblock \emph{International Journal of Engineering Research and Applications}, 2\penalty0 (4):\penalty0 693--697, 2012.

\bibitem[De~Caigny et~al.(2018)De~Caigny, Coussement, and De~Bock]{de2018new}
Arno De~Caigny, Kristof Coussement, and Koen~W De~Bock.
\newblock A new hybrid classification algorithm for customer churn prediction based on logistic regression and decision trees.
\newblock \emph{European Journal of Operational Research}, 269\penalty0 (2):\penalty0 760--772, 2018.

\bibitem[Jain et~al.(2020)Jain, Khunteta, and Srivastava]{jain2020churn}
Hemlata Jain, Ajay Khunteta, and Sumit Srivastava.
\newblock Churn prediction in telecommunication using logistic regression and logit boost.
\newblock \emph{Procedia Computer Science}, 167:\penalty0 101--112, 2020.

\bibitem[Hur and Lim(2005)]{hur2005customer}
Yeon Hur and Sehun Lim.
\newblock Customer churning prediction using support vector machines in online auto insurance service.
\newblock In \emph{International Symposium on Neural Networks}, pages 928--933. Springer, 2005.

\bibitem[Ahn et~al.(2020)Ahn, Hwang, Kim, Choi, and Kang]{ahn2020survey}
Jaehyun Ahn, Junsik Hwang, Doyoung Kim, Hyukgeun Choi, and Shinjin Kang.
\newblock A survey on churn analysis in various business domains.
\newblock \emph{IEEE Access}, 8:\penalty0 220816--220839, 2020.

\bibitem[Kumar and Chandrakala(2017)]{kumar2017optimal}
A~Saran Kumar and D~Chandrakala.
\newblock An optimal churn prediction model using support vector machine with adaboost.
\newblock \emph{Int. J. Sci. Res. Comput. Sci. Eng. Inf. Technol}, 2\penalty0 (1):\penalty0 225--230, 2017.

\bibitem[Dong et~al.(2020)Dong, Yu, Cao, Shi, and Ma]{dong2020survey}
Xibin Dong, Zhiwen Yu, Wenming Cao, Yifan Shi, and Qianli Ma.
\newblock A survey on ensemble learning.
\newblock \emph{Frontiers of Computer Science}, 14\penalty0 (2):\penalty0 241--258, 2020.

\bibitem[Hu(2005)]{hu2005data}
Xiaohua Hu.
\newblock A data mining approach for retailing bank customer attrition analysis.
\newblock \emph{Applied Intelligence}, 22\penalty0 (1):\penalty0 47--60, 2005.

\bibitem[Jinbo et~al.(2007)Jinbo, Xiu, and Wenhuang]{jinbo2007application}
Shao Jinbo, Li~Xiu, and Liu Wenhuang.
\newblock The application ofadaboost in customer churn prediction.
\newblock In \emph{2007 International Conference on Service Systems and Service Management}, pages 1--6. IEEE, 2007.

\bibitem[Kumar et~al.(2008)Kumar, Ravi, et~al.]{kumar2008predicting}
Dudyala~Anil Kumar, Vadlamani Ravi, et~al.
\newblock Predicting credit card customer churn in banks using data mining.
\newblock \emph{Int. J. Data Anal. Tech. Strateg.}, 1\penalty0 (1):\penalty0 4--28, 2008.

\bibitem[Pamina et~al.(2019)Pamina, Raja, SathyaBama, Sruthi, VJ, et~al.]{pamina2019effective}
Jeyakumar Pamina, Beschi Raja, S~SathyaBama, MS~Sruthi, Aiswaryadevi VJ, et~al.
\newblock An effective classifier for predicting churn in telecommunication.
\newblock \emph{Jour of Adv Research in Dynamical \& Control Systems}, 11, 2019.

\bibitem[Ullah et~al.(2019)Ullah, Raza, Malik, Imran, Islam, and Kim]{ullah2019churn}
Irfan Ullah, Basit Raza, Ahmad~Kamran Malik, Muhammad Imran, Saif~Ul Islam, and Sung~Won Kim.
\newblock A churn prediction model using random forest: analysis of machine learning techniques for churn prediction and factor identification in telecom sector.
\newblock \emph{IEEE access}, 7:\penalty0 60134--60149, 2019.

\bibitem[Xie et~al.(2009)Xie, Li, Ngai, and Ying]{xie2009customer}
Yaya Xie, Xiu Li, EWT Ngai, and Weiyun Ying.
\newblock Customer churn prediction using improved balanced random forests.
\newblock \emph{Expert Systems with Applications}, 36\penalty0 (3):\penalty0 5445--5449, 2009.

\bibitem[Tang et~al.(2020)Tang, Xia, Zhang, and Long]{tang2020customer}
Qi~Tang, Guoen Xia, Xianquan Zhang, and Feng Long.
\newblock A customer churn prediction model based on xgboost and mlp.
\newblock In \emph{2020 International Conference on Computer Engineering and Application (ICCEA)}, pages 608--612. IEEE, 2020.

\bibitem[{\c{C}}elik and Osmanoglu(2019)]{ccelik2019comparing}
Ozer {\c{C}}elik and Usame~O Osmanoglu.
\newblock Comparing to techniques used in customer churn analysis.
\newblock \emph{Journal of Multidisciplinary Developments}, 4\penalty0 (1):\penalty0 30--38, 2019.

\bibitem[Domingos et~al.(2021)Domingos, Ojeme, and Daramola]{domingos2021experimental}
Edvaldo Domingos, Blessing Ojeme, and Olawande Daramola.
\newblock Experimental analysis of hyperparameters for deep learning-based churn prediction in the banking sector.
\newblock \emph{Computation}, 9\penalty0 (3):\penalty0 34, 2021.

\bibitem[Cenggoro et~al.(2021)Cenggoro, Wirastari, Rudianto, Mohadi, Ratj, and Pardamean]{cenggoro2021deep}
Tjeng~Wawan Cenggoro, Raditya~Ayu Wirastari, Edy Rudianto, Mochamad~Ilham Mohadi, Dinne Ratj, and Bens Pardamean.
\newblock Deep learning as a vector embedding model for customer churn.
\newblock \emph{Procedia Computer Science}, 179:\penalty0 624--631, 2021.

\bibitem[Panjasuchat and Limpiyakorn(2020)]{panjasuchat2020applying}
M~Panjasuchat and Y~Limpiyakorn.
\newblock Applying reinforcement learning for customer churn prediction.
\newblock In \emph{Journal of Physics: Conference Series}, volume 1619, page 012016. IOP Publishing, 2020.

\bibitem[Cao et~al.(2019)Cao, Liu, Chen, and Zhu]{cao2019deep}
Shulin Cao, Wei Liu, Yuxing Chen, and Xiaoyan Zhu.
\newblock Deep learning based customer churn analysis.
\newblock In \emph{2019 11th International Conference on Wireless Communications and Signal Processing (WCSP)}, pages 1--6. IEEE, 2019.

\bibitem[Wangperawong et~al.(2016)Wangperawong, Brun, Laudy, and Pavasuthipaisit]{wangperawong2016churn}
Artit Wangperawong, Cyrille Brun, Olav Laudy, and Rujikorn Pavasuthipaisit.
\newblock Churn analysis using deep convolutional neural networks and autoencoders.
\newblock \emph{arXiv preprint arXiv:1604.05377}, 2016.

\bibitem[Fujo et~al.(2022)Fujo, Subramanian, Khder, et~al.]{fujo2022customer}
Samah~Wael Fujo, Suresh Subramanian, Moaiad~Ahmad Khder, et~al.
\newblock Customer churn prediction in telecommunication industry using deep learning.
\newblock \emph{Information Sciences Letters}, 11\penalty0 (1):\penalty0 24, 2022.

\bibitem[Umayaparvathi and Iyakutti(2017)]{umayaparvathi2017automated}
V~Umayaparvathi and K~Iyakutti.
\newblock Automated feature selection and churn prediction using deep learning models.
\newblock \emph{International Research Journal of Engineering and Technology (IRJET)}, 4\penalty0 (3):\penalty0 1846--1854, 2017.

\bibitem[Krizhevsky et~al.(2012)Krizhevsky, Sutskever, and Hinton]{krizhevsky2012imagenet}
Alex Krizhevsky, Ilya Sutskever, and Geoffrey~E Hinton.
\newblock Imagenet classification with deep convolutional neural networks.
\newblock \emph{Advances in neural information processing systems}, 25, 2012.

\bibitem[Szegedy et~al.(2015)Szegedy, Liu, Jia, Sermanet, Reed, Anguelov, Erhan, Vanhoucke, and Rabinovich]{szegedy2015going}
Christian Szegedy, Wei Liu, Yangqing Jia, Pierre Sermanet, Scott Reed, Dragomir Anguelov, Dumitru Erhan, Vincent Vanhoucke, and Andrew Rabinovich.
\newblock Going deeper with convolutions.
\newblock In \emph{Proceedings of the IEEE conference on computer vision and pattern recognition}, pages 1--9, 2015.

\bibitem[Szegedy et~al.(2016)Szegedy, Vanhoucke, Ioffe, Shlens, and Wojna]{szegedy2016rethinking}
Christian Szegedy, Vincent Vanhoucke, Sergey Ioffe, Jon Shlens, and Zbigniew Wojna.
\newblock Rethinking the inception architecture for computer vision.
\newblock In \emph{Proceedings of the IEEE conference on computer vision and pattern recognition}, pages 2818--2826, 2016.

\bibitem[Szegedy et~al.(2017)Szegedy, Ioffe, Vanhoucke, and Alemi]{szegedy2017inception}
Christian Szegedy, Sergey Ioffe, Vincent Vanhoucke, and Alexander~A Alemi.
\newblock Inception-v4, inception-resnet and the impact of residual connections on learning.
\newblock In \emph{Thirty-first AAAI conference on artificial intelligence}, 2017.

\bibitem[Li et~al.(2022)Li, Wang, Huang, and Zhou]{li2022convnext}
Jiachen Li, Chixin Wang, Banban Huang, and Zekun Zhou.
\newblock Convnext-backbone hovernet for nuclei segmentation and classification.
\newblock \emph{arXiv preprint arXiv:2202.13560}, 2022.

\bibitem[Vaswani et~al.(2017)Vaswani, Shazeer, Parmar, Uszkoreit, Jones, Gomez, Kaiser, and Polosukhin]{vaswani2017attention}
Ashish Vaswani, Noam Shazeer, Niki Parmar, Jakob Uszkoreit, Llion Jones, Aidan~N Gomez, {\L}ukasz Kaiser, and Illia Polosukhin.
\newblock Attention is all you need.
\newblock \emph{Advances in neural information processing systems}, 30, 2017.

\bibitem[Liu et~al.(2021)Liu, Lin, Cao, Hu, Wei, Zhang, Lin, and Guo]{liu2021swin}
Ze~Liu, Yutong Lin, Yue Cao, Han Hu, Yixuan Wei, Zheng Zhang, Stephen Lin, and Baining Guo.
\newblock Swin transformer: Hierarchical vision transformer using shifted windows.
\newblock In \emph{Proceedings of the IEEE/CVF International Conference on Computer Vision}, pages 10012--10022, 2021.

\bibitem[Simonyan and Zisserman(2014)]{simonyan2014very}
Karen Simonyan and Andrew Zisserman.
\newblock Very deep convolutional networks for large-scale image recognition.
\newblock \emph{arXiv preprint arXiv:1409.1556}, 2014.

\bibitem[He et~al.(2016)He, Zhang, Ren, and Sun]{he2016deep}
Kaiming He, Xiangyu Zhang, Shaoqing Ren, and Jian Sun.
\newblock Deep residual learning for image recognition.
\newblock In \emph{Proceedings of the IEEE conference on computer vision and pattern recognition}, pages 770--778, 2016.

\bibitem[Cui et~al.(2016)Cui, Chen, and Chen]{cui2016multi}
Zhicheng Cui, Wenlin Chen, and Yixin Chen.
\newblock Multi-scale convolutional neural networks for time series classification.
\newblock \emph{arXiv preprint arXiv:1603.06995}, 2016.

\bibitem[Chen(2015)]{chen2015convolutional}
Yahui Chen.
\newblock Convolutional neural network for sentence classification.
\newblock Master's thesis, University of Waterloo, 2015.

\bibitem[Ismail~Fawaz et~al.(2019)Ismail~Fawaz, Forestier, Weber, Idoumghar, and Muller]{ismail2019deep}
Hassan Ismail~Fawaz, Germain Forestier, Jonathan Weber, Lhassane Idoumghar, and Pierre-Alain Muller.
\newblock Deep learning for time series classification: a review.
\newblock \emph{Data mining and knowledge discovery}, 33\penalty0 (4):\penalty0 917--963, 2019.

\bibitem[Hoseinzade and Haratizadeh(2019)]{hoseinzade2019cnnpred}
Ehsan Hoseinzade and Saman Haratizadeh.
\newblock Cnnpred: Cnn-based stock market prediction using a diverse set of variables.
\newblock \emph{Expert Systems with Applications}, 129:\penalty0 273--285, 2019.

\bibitem[Weyn et~al.(2020)Weyn, Durran, and Caruana]{weyn2020improving}
Jonathan~A Weyn, Dale~R Durran, and Rich Caruana.
\newblock Improving data-driven global weather prediction using deep convolutional neural networks on a cubed sphere.
\newblock \emph{Journal of Advances in Modeling Earth Systems}, 12\penalty0 (9):\penalty0 e2020MS002109, 2020.

\bibitem[Dobko et~al.(2020)Dobko, Petryshak, and Dobosevych]{dobko2020cnn}
Mariia Dobko, Bohdan Petryshak, and Oles Dobosevych.
\newblock Cnn-cass: Cnn for classification of coronary artery stenosis score in mpr images.
\newblock \emph{arXiv preprint arXiv:2001.08593}, 2020.

\bibitem[Ismail~Fawaz et~al.(2020)Ismail~Fawaz, Lucas, Forestier, Pelletier, Schmidt, Weber, Webb, Idoumghar, Muller, and Petitjean]{ismail2020inceptiontime}
Hassan Ismail~Fawaz, Benjamin Lucas, Germain Forestier, Charlotte Pelletier, Daniel~F Schmidt, Jonathan Weber, Geoffrey~I Webb, Lhassane Idoumghar, Pierre-Alain Muller, and Fran{\c{c}}ois Petitjean.
\newblock Inceptiontime: Finding alexnet for time series classification.
\newblock \emph{Data Mining and Knowledge Discovery}, 34\penalty0 (6):\penalty0 1936--1962, 2020.

\bibitem[Liu et~al.(2022)Liu, Mao, Wu, Feichtenhofer, Darrell, and Xie]{liu2022convnet}
Zhuang Liu, Hanzi Mao, Chao-Yuan Wu, Christoph Feichtenhofer, Trevor Darrell, and Saining Xie.
\newblock A convnet for the 2020s.
\newblock In \emph{Proceedings of the IEEE/CVF Conference on Computer Vision and Pattern Recognition}, pages 11976--11986, 2022.

\bibitem[Dosovitskiy et~al.(2020)Dosovitskiy, Beyer, Kolesnikov, Weissenborn, Zhai, Unterthiner, Dehghani, Minderer, Heigold, Gelly, et~al.]{dosovitskiy2020image}
Alexey Dosovitskiy, Lucas Beyer, Alexander Kolesnikov, Dirk Weissenborn, Xiaohua Zhai, Thomas Unterthiner, Mostafa Dehghani, Matthias Minderer, Georg Heigold, Sylvain Gelly, et~al.
\newblock An image is worth 16x16 words: Transformers for image recognition at scale.
\newblock \emph{arXiv preprint arXiv:2010.11929}, 2020.

\bibitem[Kingma and Ba(2014)]{kingma2014adam}
Diederik~P Kingma and Jimmy Ba.
\newblock Adam: A method for stochastic optimization.
\newblock \emph{arXiv preprint arXiv:1412.6980}, 2014.

\bibitem[Ioffe and Szegedy(2015)]{ioffe2015batch}
Sergey Ioffe and Christian Szegedy.
\newblock Batch normalization: Accelerating deep network training by reducing internal covariate shift.
\newblock In \emph{International conference on machine learning}, pages 448--456. PMLR, 2015.

\bibitem[Srivastava et~al.(2014)Srivastava, Hinton, Krizhevsky, Sutskever, and Salakhutdinov]{srivastava2014dropout}
Nitish Srivastava, Geoffrey Hinton, Alex Krizhevsky, Ilya Sutskever, and Ruslan Salakhutdinov.
\newblock Dropout: a simple way to prevent neural networks from overfitting.
\newblock \emph{The journal of machine learning research}, 15\penalty0 (1):\penalty0 1929--1958, 2014.

\bibitem[Sergeev and Balso(2018)]{sergeev2018horovod}
Alexander Sergeev and Mike~Del Balso.
\newblock Horovod: fast and easy distributed deep learning in {TensorFlow}.
\newblock \emph{arXiv preprint arXiv:1802.05799}, 2018.

\end{thebibliography}






\end{document}